%% file: main.tex
\documentclass[10pt,twocolumn,letterpaper]{article}

\usepackage{cvpr}
\usepackage{times}
\usepackage[utf8]{inputenc} 
\usepackage[T1]{fontenc}    
\usepackage{url}            
\usepackage{booktabs}       
\usepackage{amsfonts}       
\usepackage{amsmath}        
\usepackage{amssymb}        
\usepackage{nicefrac}       
\usepackage{microtype}      
\usepackage[dvipsnames]{xcolor} 
\usepackage{graphicx} 
\graphicspath{{images/}}
\DeclareGraphicsExtensions{.pdf,.jpeg,.png}
\usepackage[caption=false]{subfig}  
\usepackage{multicol}       
\usepackage{multirow}       
\usepackage{comment}        
\usepackage{array}
\newcolumntype{C}[1]{>{\centering\arraybackslash}p{#1}}

\usepackage[nolist]{acronym}    
\usepackage[toc,page]{appendix} 

\newcommand{\note}[2]{%
	\fbox{\textcolor{red}{\bfseries\sffamily\scriptsize#1}}%
	\ifx&#2&%
	\else
	  {\sffamily\small$\triangleright$\textit{#2}$\triangleleft$}
	\fi
}

\usepackage[breaklinks=true,bookmarks=false]{hyperref}

\cvprfinalcopy 

\def\cvprPaperID{7354} 

\ifcvprfinal\fi
\begin{document}

\title{Trainable Spectrally Initializable Matrix Transformations in \\ Convolutional Neural Networks}

\author{%
    \textbf{Michele~Alberti\thanks{Equal contribution}$^{\hphantom{*},1,5}$ \enskip Angela~Botros$^{*,2}$  \enskip Narayan~Sch\"utz$^{*,2}$}\\ 
    \textbf{Rolf~Ingold$^1$  \enskip Marcus~Liwicki$^3$  \enskip Mathias~Seuret$^{1,4}$} \\
  $^1$Document Image and Voice Analysis Group (DIVA), University of Fribourg, Switzerland\\
  $^2$ARTORG Center for Biomedical Engineering Research, University of Bern, Switzerland\\
  $^3$EISLAB Machine Learning, Lule{\aa} University of Technology, Sweden\\
  $^4$Pattern Recognition Lab, Friedrich-Alexander-Universit\"at Erlangen-N\"urnberg, Germany\\
  $^5$V7 Ltd, London, United Kingdom
}

\maketitle
\thispagestyle{empty}

\input{sections/acronym.tex}

\input{sections/abstract}

\input{sections/introduction.tex}

\input{sections/related_work.tex}

\input{sections/spectrallayer.tex}
\input{sections/experimental_setup.tex}

\input{sections/results.tex}

\input{sections/conclusion.tex}


\section*{Acknowledgment}
The work in this paper has been partially supported by the HisDoc III project funded by the Swiss National Science Foundation with the grant number $205120$\textunderscore$169618$.

\clearpage 

{\small
\bibliographystyle{ieee_fullname}
\bibliography{biblio}
}

\clearpage 
\input{sections/appendices.tex}

\end{document}

%% file: sections/acronym.tex
\begin{acronym}

\acro{Abb.}{Abbrevation}

\acro{fxd}{Fixed Layer}
\acro{ufx}{Unfixed Layer}
\acro{mxd}{Mixed Layer}
\acro{sp}{Spectral Layer}
\acro{fc}{Fully Connected Linear Layer}

\acro{CNN}{Convolutional Neural Network}
\acro{MLP}{Multi Layer Perceptron}
\acro{ReLU}{Rectified Linear Units}
\acro{NN}{Neural Network}

\acro{DFT}{Discrete Fourier Transformation}
\acro{iDFT}{Inverse Discrete Fourier Transformation}
\acro{DCT}{Discrete Cosine Transformation}
\acro{iDCT}{Inverse Discrete Cosine Tranformation}

\acro{GPU}{Graphics Processor Units}

\acro{LDA}{Linear Discriminant Analysis}
\acro{PCA}{Principal Component Analysis}

\acro{MRI}{Magnetic Resonance Imaging}
\acro{FTIR}{Fourier Transformed Infrared Spectrum}

\acro{IRMAS}{Instrument Recognition in Musical and Audio Signals}
\acro{CIFAR-10}{Canadian Institute For Advanced Research}
\acro{MNIST}{Modified National Institute of Standards and Technology}
\acro{FashionMNIST}{Fashion Modified National Institute of Standards and Technology}
\acro{ColorectalHist}{Colorectal Cancer Histology}

\acro{MLP}{Multilayer Perceptron}
\acro{GAP}{Global Average Pooling}

\end{acronym}

%% file: sections/abstract.tex
\begin{abstract}


In this work, we investigate the application of trainable and spectrally initializable matrix transformations on the feature maps produced by convolution operations.
While previous literature has already demonstrated the possibility of adding static spectral transformations as feature processors, our focus is on more general trainable transforms.
We study the transforms in various architectural configurations on four datasets of different nature: from medical (ColorectalHist, HAM10000) and natural (Flowers, ImageNet) images to historical documents (CB55) and handwriting recognition (GPDS).
With rigorous experiments that control for the number of parameters and randomness, we show that networks utilizing the introduced matrix transformations outperform vanilla neural networks. 
The observed accuracy increases by an average of 2.2\,\% across all datasets.
In addition, we show that the benefit of spectral initialization leads to significantly faster convergence, as opposed to randomly initialized matrix transformations. 
The transformations are implemented as auto-differentiable PyTorch modules that can be incorporated into any neural network architecture.
The entire code base is open-source.
\end{abstract}

%% file: sections/introduction.tex
\section{Introduction}
\label{toc:introduction}

In recent years neural networks experienced a renaissance, leading to numerous performance breakthroughs in many machine learning tasks~\cite{schmidhuber2015deep, lecun2015deep}. 
Besides the simpler \ac{MLP}~\cite{mlp, schmidhuber2015deep}, especially \ac{CNN}~\cite{fukushima1980neocognitron, lecuncnn} were increasingly popularized.
The progressively deeper architectures of neural networks allowed them to successfully learn hierarchical representations of data, in other words, to implicitly learn suitable feature representations.
This, in turn, makes it feasible to train on raw data in an end-to-end fashion, as is currently the state of the art in image recognition~\cite{lecun2015deep}. 
\ac{CNN} generally operate on locally distinct parts of the input. This localization invariant behavior can be highly beneficial to network performance.
As a disadvantage, vanilla \acp{CNN} are not able to apply transformations that concern the global image (or feature map) structure e.g., translation, rotation, scaling, mirroring and shearing.
These transformations are common in a variety of applications such as visual computing and signal processing.
Prominent examples of such transformations are unitary spectral transforms such as the \acf{DFT} or \acf{DCT}.
Depending on the input domain, such transforms can lead to better representations of the data, making the task at hand substantially easier (faster convergence and/or higher accuracy).
So far, these transformations are often manually applied in a preprocessing step, before feeding the data into the CNN.
This is common practice in audio signal or structured image related learning tasks~\cite{MelSpectrogram,DeepFeatureDCT,MixedFeatureNeuralNetwork}.


In this work, we introduce trainable matrix transformations on images or feature maps, based on the structure of the DCT and DFT.
The idea here is similar in nature to the spatial transform networks~\cite{Jaderberg2015spatialtransformer} but instead of input conditioned spatial transforms we explore the possibility of learning a more general matrix transformation that can be initialized as a pre-specified unitary transform (DCT and DFT) and is applicable to the whole input domain of a respective layer.
The rationale for initializing in the spectral domain is based on the observation that it often leads to more compact feature representations, resulting in faster convergence~\cite{SpectralConvolutionalNeuralNetworks}.
The main advantage of having such trainable transforms over a fixed preprocessing based image transformation is that it requires less expert knowledge and can thus adjust automatically to a given input domain, effectively allowing for end-to-end learning. 

\subsection*{Contribution}
\label{toc:contribution}

The main contribution of this paper is two-fold. 
First, we implement trainable linear matrix transformation modules, which can be used as a priori initialized spectral transforms (\ac{DCT}, \ac{DFT}) or the random normalized initialization~\cite{glorot2010understanding}.
All transforms are differentiable and can be trained with regular gradient descent based algorithms and are thus straight forward to incorporate into regular CNNs in any major framework. 
Second, to evaluate the usefulness of the transforms in \ac{CNN}s, we rigorously evaluated classification performance on four publicly available 2D datasets.
To the best of our knowledge, we are the first to develop and integrate trainable linear transforms in the proposed way.
Our PyTorch based  open source implementations are freely available as a pip installable python package\footnote{\url{https://github.com/NarayanSchuetz/SpectralLayersPyTorch}} and have already been integrated\footnote{\url{https://github.com/NarayanSchuetz/DeepDIVA}} into the DeepDIVA~\cite{albertipondenkandath2018deepdiva} deep learning framework thus enabling full reproducibility of experiments.

%% file: sections/related_work.tex
\section{Related Work}
\label{toc:related_Work}

The structure of the spectral domain has been exploited for many feature extraction purposes.
Pictures of tissue samples, cancer structure or \acf{MRI} profit from features obtained through spectral transformations.
In~\cite{MixedFeatureNeuralNetwork} the authors have shown that mammographies can be classified well by using features from original and \acs{DCT}-transformed pictures.
\acf{FTIR} of relevant tissue samples for cancer detection has been around since the 90s~\cite{meurens1996breast,sukuta1999factor}.
Currently, integration of \acs{FTIR} in classification tasks provides better results in cancer detection~\cite{gajjar2013fourier}.
In~\cite{BrainTumorDCT}, \acs{DCT} is used both for dimensionality reduction of brain tumor \acs{MRI}s, as well as for feature extraction, serving as input to a \ac{NN}.
Regular or irregular patterns of different sizes, i.e., the tissue structure and other medically relevant details, are well suited for analysis with spectral methods. 
This is opposed to recognition of large objects, e.g. faces, cars or people, where usage of \acp{CNN} is well established~\cite{krizhevsky2012,he2015,hu2017}.

Work from Rippel et al.\ investigates the use of \acs{DFT} in neural networks~\cite{SpectralConvolutionalNeuralNetworks}.
They introduce the concepts of spectral pooling and spectral parametrization.
Spectral pooling is comparable to low-pass filtering the data and has a similar effect on the number of parameters as \emph{max-pooling} on the original image.
In spectral parametrization, convolutional filters are trained in the spectral domain.
After training, the filters are transformed back to the spatial domain.
This method generates the same filters as obtained through regular training, but the convergence rate is much higher.
In~\cite{DeepFeatureDCT} they showed that static \acs{DCT} transforms applied on feature maps led to better convergence rates on different data sets. 
In~\cite{Kovacs2013spectrotemporal}, a trainable \acs{DCT} layer is implemented to match the classic \ac{MLP} layers and showed similar results as a Gabor filter layer in speech recognition.
\cite{Jaderberg2015spatialtransformer} introduces the idea to apply parameterized spatial transformations on feature maps. 
In~\cite{anden2014deep} and~\cite{variani2016complex} scattering transforms are used to generate more stable representation of data in a theoretically well founded but practically more complex manner.

Reducing the complexity of linear layers has been tackled by various groups~\cite{sindhwani_structured_2015, moczulski_acdc:_2016}.
In~\cite{sindhwani_structured_2015}, the structure of special matrix layers, such as Vandermonde, Toepliz or Cauchy matrices, is exploited to reduce memory space.
In~\cite{moczulski_acdc:_2016}, so called structured efficient linear layers (SELLs), a combination of diagonal, sparse or permutation matrices, and implementations of Fourier, Hadamard and Cosine transformations are investigated.
They show that the usage of these SELLs allowes to approximate dense layers while reducing memory usage and complexity of the \ac{NN}.




The idea to initialize weights in \ac{NN} according to prior assumptions or information has proven successful time and time again. This can be seen in commonly used transfer learning applications~\cite{pan2010survey} but also with more specific initialization of theoretically well-founded linear functions like Gabor filters~\cite{gabor}, \acf{PCA}~\cite{seuret2017pca} and \ac{LDA}~\cite{alberti2017historical}.
In the latter, they pushed initialization further by using label information to directly produce networks with classification abilities.

%% file: sections/spectrallayer.tex
\section{Matrix Transforms in Neural Networks}
\label{toc:SpectralLayers}
In this section, the mathematical background of the implemented linear mappings is explained in detail. \newline
For two vector spaces $U$ and $V$, a function $f: U \rightarrow V$ is a linear mapping if for $x,y \in U$ and any scalar $c$ it holds that $f(cx+cy) = cf(x)+cf(y)$.

In this paper, we look at layers of \ac{NN}, implemented in the form of two matrix multiplications. 
They can be described by a linear mapping 

\begin{align}
g&: U \rightarrow V, & U\!\subseteq\! \mathbb{R}^{N\times M}, V\!\subseteq \! \mathbb{R}^{K\times L} \label{linmap1}\\
g(x)&: x \mapsto W_1\!\cdot\! x\! \cdot\! W_2^T, & W_1\! \in\! \mathbb{R}^{K\times N}, W_2\! \in\! \mathbb{R}^{L\times M}\label{linmap2}
\end{align}
This is a composition of two linear mappings $g(x) = g_2 \circ g_1 (x)$, with $g_1: x \mapsto W_1 \cdot x$ and $g_2: \tilde{x} \mapsto \tilde{x}\cdot W_2^T$.
As a side note, the linear mapping as described in~\eqref{linmap2} can be reformulated using the Kronecker product.
While this is a convenient approach when solving linear systems, it is not advisable for the implementation in a neural network. 
The resulting block matrices would have more parameters than the original ones which would unnecessarily increase computational complexity.
More details on the Kronecker product and why we chose no to use it are provided in appendix~\ref{app:kronecker}
Besides this general linear mapping, in this paper we look at two specific mappings, the \acf{DFT} and the \acf{DCT}.
These are commonly considered as spectral transformations, as they transform a spatial representation to a spectral (frequency) representation.

For a real-valued image $x \in \mathbb{R}^{N\times M}$, the \acf{DFT} can be written as: 
\begin{align}
    X &= W_K \cdot x \cdot W_L^T,\label{fourier2DM_1} \\
    W_K[k,n] &= e^{-2\pi j \frac{k\cdot n}{N}},~~ W_L[l,m] = e^{-2\pi j \frac{l\cdot m}{M}} \label{fourier2DM}
\end{align}
for $k = [0 ... K-1],~ n = [0 ... N-1],~l = [0 ... L-1] $ and $m = [0 ... M-1]$.
The 2D-DFT of $x$ is $X~\in~\mathbb{C}^{K\times L}$. The variables $W_K~\in~\mathbb{C}^{K \times N}$ and $W_L~\in~\mathbb{C}^{L\times M}$ are the Fourier transformation matrices. 
The parameters $K$ and $L$ denote the number of vertical and horizontal frequency components by which the input image is represented. 
Both $W_K$ and $W_L$ are unitary matrices, making the two matrix multiplications unitary transformations, i.e., the matrix is only rotated and possibly mirrored, but not scaled.
While the DFT matrices have Vandermonde structure, this will not be exploited, in order to avoid complex parameters and operations.
To ensure real parameters, the real and imaginary parts are separated and computed individually using Euler's formula
$e^{jx} = \cos(x) + j \cdot \sin(x) \label{euler}$.
In our implementation, every complex parameter is treated as a real-valued vector $v = [\Re(v),~~ \Im(v)]^T$ with $\Re(\cdot)$ denoting the real part and $\Im(\cdot)$ the imaginary part. 
This doubles the number of parameters for the \ac{DFT} implementation. 
More details for the implementation are given in the supplementary material in Appendix~\ref{app:layerImplementation}.
For a real-valued input signal, the Fourier transform has important symmetry properties. 
These and their implications for the transformation are discussed in~\cite{SpectralConvolutionalNeuralNetworks} and in Appendix~\ref{app:redundancy}.
The back transformation of the \ac{DFT} is implemented in the same manner as the forward transformation~\cite{DCT_DFT}.
More details on it are given in Appendix~\ref{app:backtransform}.

The 2D-DCT II\footnote{There are four common implementations of the DCT. And while all have advantages and disadvantages, the DCT II is the most commonly used~\cite{DCT_DFT}.} of an image $x\in \mathbb{R}^{N\times M}$ can be written as:
\begin{align}
    X & = W_K\! \cdot\! x\! \cdot\! W_L^T, \label{dct2DM_1}\\ 
    &\begin{array}{l}
        W_K[k,n]   = \cos{\!\left(\!\frac{\pi}{N}\!\left(\!n + \frac12 \!\right)k\right)}\\
        W_L[l,m]   = \cos{\!\left(\!\frac{\pi}{M}\!\left(\!m + \frac12 \!\right)l\right)}
    \end{array} \label{dct2DM}
\end{align}
for $k = [0 ... K-1],~ n = [0 ... N-1], l = [0 ... L-1] $ and $m = [0 ... M-1]$.
This transformation is similar to the \acs{DFT}, but the resulting signal is real. 
The two variables $W_K~\in~\mathbb{R}^{K \times N}$ and $W_L~\in~\mathbb{R}^{L \times M}$ are the orthogonal cosine transformation matrices, i.e., the matrix is rotated and possibly mirrored, but not scaled. 
As with the \ac{DFT}, the parameters $K$ and $L$ denote the number of frequencies used in the transformation to represent the input image in the frequency space.
Again, the back transformation is implemented in the same manner as the forward transformation.
More details on it are in Appendix~\ref{app:backtransform}.

Both the 2D-\ac{DFT} and the 2D-\ac{DCT} are linear mapings as defined in~\eqref{linmap1} and~\eqref{linmap2},
mapping from $U = \mathbb{R}^{N\times M}$ to $V = \mathbb{R}^{K\times L}$ (DCT) or $V = \mathbb{C}^{K\times L}$ (DFT) respectively. 
As they are linear mappings, the gradient of this composition of matrix-matrix multiplications is the product of the transformation matrices~\cite{SpectralConvolutionalNeuralNetworks}.
\begin{align}
    x \mapsto W_K  x  W_L^T \!=\! X
    \! \Leftrightarrow \!
    \left\{\!\!\! \begin{array}{l}
    X_1 \!=\! W_K \cdot x \\
    X \!=\! X_1W_L^T \!=\! (W_LX_1^T)^T
    \end{array}
    \right.\label{linearlayer}
\end{align}
There is a strong similarity between this linear mapping as depicted in~\eqref{linearlayer} and the classic linear layer of the \ac{MLP} that is defined for a vector input $v \in \mathbb{R}^N$ and is given as $v\mapsto f(W\cdot v)$ with $f(\cdot)$ being the activation function.
A short overview over the similarities and differences is given in Appendix~\ref{app:matrixdimensions}.

%% file: sections/experimental_setup.tex
\section{Experimental Setting}
\label{toc:experimental_setting}

\begin{figure*}
    \centering
    \includegraphics[width=\textwidth]{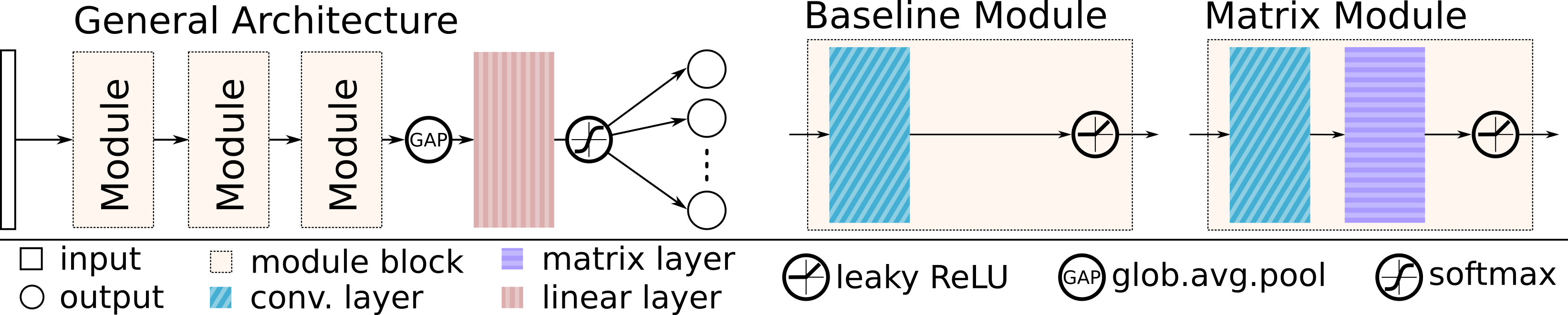}
    \caption{
    Depicted on the left is the general network architecture used for the image classification task, in the middle the baseline module structure and on the right the spectral module structure.
    Section~\ref{toc:2D_architecture} explains each module and architecture in detail. 
    The spectral part (purple) can be \ac{DFT} initialized, \ac{DCT} initialized (including the respective inverse transform) or random initialized of same size and structure. 
    Section~\ref{toc:models_configuration} describes the different module configurations used in this paper.
}
    \label{fig:model_architecture}
\end{figure*}

\subsection{Task and Data}
\label{toc:task_datasets}

We consider the image classification task. 
Given an input, we produce a single output label which corresponds to what is contained in the image.
To asses the generality of our findings, we perform this on images from different source domains, i.e., images from histologic sections, dermatoscopic images, high-resolution scans of historical documents and natural images.
The dataset decision is a deliberate choice.
We believe that to prove the effectiveness of our method it is necessary to study realistic problems.
Therefore, the chosen datasets include images from sensible domains such as medical imaging where the application of spectral analysis has proven useful in the past, and signatures, where is it common to find dataset of limited size due to the difficulties in collecting and annotating large amount of data.
Additionally, the choice is steered by the strengths of the matrix-layer approach, which include its ability to handle large input dimensions with a reasonable amount of trainable parameters (see Appendix \ref{app:matrixdimensions}).
Specifically, we use the following public datasets.

The Colorectal Cancer Histology (ColorectalHist) dataset~\cite{colorectal} is an image collection of $5K$ specimens from histologic sections that depict eight mutually exclusive tissue types. 

The Human Against Machine (HAM10000) dataset~\cite{HAM10000} is composed of $10K$ pigmented skin lesion images, which include a representative collection of all (seven) important diagnostic categories in the realm of pigmented lesions.

The CB55 is a manuscript extract from the DIVA-HisDB dataset~\cite{simistira_2016_diva} which consists of challenging medieval manuscripts, precisely annotated for the evaluation of several Document Image Analysis (DIA) tasks. 
The task consist into distinguishing the four classes: main text, comment, decoration and background.
Given the intractable large input size of every page (high-resolution scans at 24MP), in this work we selected $100K$ square patches of size $149 \times 149$ and assigned them the label of their central pixel, thus creating a classification dataset.

The Flowers Recognition dataset~\cite{flowers} contains $4.2K$ images of five different types of flowers.
The data collection is based on Flickr, Google images and Yandex images. 

The ImageNet dataset~\cite{deng2009imagenet} is a well-known dataset used in several computer vision tasks.
In this work we used the Large Scale Visual Recognition Challenge 2016 (ILSVRC2016) subset which contains object belonging to $1K$ different categories and is composed of $1.2M$ images.

The GPDSsynthetic-Offline (GPDS) is a large synthetic dataset~\cite{Ferrer2015} that contains $4K$ synthetic users with $24$~genuine signatures and 30~simulated forgeries each. 
The simulated resolution of the images is $600$ dpi.
In this work we employ the GPDS-last100 subset as described in~\cite{maergner2019combining}.

\subsection{Convolutional Architecture}
\label{toc:2D_architecture}

In order to operate on image-based datasets, we designed a \ac{CNN} model, shown in Figure~\ref{fig:model_architecture}.
The core structure consists of 3 module-block layers, followed by a \ac{GAP} and concluded by a fully connected classification layer.
There are two types of modules: the baseline and the matrix transformation ones.
The baseline module is composed of a convolution layer followed by a Leaky ReLU activation function and is shown in Figure~\ref{fig:model_architecture} in the middle.
The matrix transformation module is similar to the baseline module with the addition of a matrix transformation right after the convolutional layer, as shown on the right of Figure~\ref{fig:model_architecture}.
This way, a linear mapping - as defined in equation~\eqref{linmap2} - is performed on the feature space and then subject to the activation function, as is common practice with neural networks. 
The receptive field of the networks do cover the whole image, thus, spectral transforms can bring benefits in regard to global structure.

In order to have comparable results, regardless of the configuration, each model has the same amount of parameters\footnote{With a $\pm3.5\%$ tolerance.} which is roughly 135K.\footnote{Details on the number of parameters in each network are provided in the appendices in Appendix~\ref{toc:appendix_number_parameters}.} 
The exact numbers are provided together with the source code, such that full reproducibility is guaranteed (details in Section~\ref{toc:reproducibility}).
Our goal is to investigate the effectiveness and applicability of general and special matrix transformations and not to set a new state-of-the-art results.
To that end, we rely on exact control with regards to the number of parameters and protocols in the respective architectures. 
In fact, the architecture is relatively small compared with state-of-the-art models as our main goal is investigating the effectiveness of matrix layers --- a comparison with state-of-the-art models is still given in the experiments. 

The reason behind our custom architecture choice is control.
Using only complex high-end architectures would harm the actual scientific methodology by introducing unfair comparison and preventing the measurement of the effect of our novel contribution in isolation.
With its only 3 layers and no additional features (such as batch normalization, skip-connection, ...) we have more control and we can thus asses whether the observed behaviour is affected by our matrix transformations initialization.
Inserting our layers into an existing - and very complex - network would hinder the reliability of causation conclusions, since the performance boost could be a byproduct of some other hidden effect.  
Finally, the experimental protocol features afixed parameter budget which is nearly impossible to achieve without altering the original ``host'' network architecture at all.

\subsection{Training Parameters and Optimization}
\label{toc:models_configuration}

In this work, we run experiments with two different training configurations (way of arranging) the modules.
One, named ``Single'', where only one matrix transform is applied (after the first convolution operation), i.e. one matrix module is followed by two baseline modules.
The other, named ``Multi'', employs multiple transforms, one after each convolutional layer, i.e. three matrix modules are used.
In both configurations the matrices are initialized with either the random normalized initialization RND~\cite{glorot2010understanding}, DFT or DCT, respectively.
For better interpretability, the ``Multi'' configurations with spectrally initialized transforms employ the respective inverse transform in the second (middle) layer.
In Table~\ref{tab:results} there is an overview of all models we use. 

In all the experiments we use Stochastic Gradient Descent (SGD) with 0.9 momentum, cross entropy loss as loss function and L2 regularization.
All other training details are provided along with the open-sourced code (Section~\ref{toc:reproducibility}).

The choice of hyper-parameters is very important, since we are comparing different initialization methods.
We use a black-box hyper-parameter optimization solution which automates model tuning to select the best values for each model we run experiments with~\cite{sigopt}.
The results reported in Table~\ref{tab:results} are selected as the mean and standard deviation of 20 independent runs, using the best hyper-parameters found after 30 optimization iterations - for each model and dataset separately.

\subsection{Reproducibility}
\label{toc:reproducibility}

In order to ensure full reproducibility of our experiments, we use the deep learning framework DeepDIVA~\cite{albertipondenkandath2018deepdiva}.
We provide a repository on GitHub (link in Section~\ref{toc:contribution}, Page 2) that contains all code necessary to reproduce our results as well as documentation in order to expand on our work.

%% file: sections/results.tex
\section{Results and Analysis}
\label{toc:results}

\begin{table*}[t]
    \caption{%
    This table reports the results of all models on the test set across all the datasets presented in Section~\ref{toc:task_datasets}.
    The metrics are mean accuracy (in \%) and respective standard deviation, computed on the basis of 20 independent runs, using the best hyper-parameters found after 30 optimization iterations - for each model and dataset separately.
    The two different spectral transformations, \ac{DFT} and \ac{DCT} are introduced in Section~\ref{toc:SpectralLayers}.
    The ``BaselineDeep'' model is introduced in Section~\ref{toc:regularization_properties}.
    The highest performance is made bold for each dataset (reference models are not taken into account as unfair comparison because of huge the difference in parameters).
    } 
    \label{tab:results}
    \begin{center}
    \begin{small}
    \begin{sc}
    \resizebox{\textwidth}{!}{%
    \begin{tabular}{c l C{4mm}C{4mm} c C{4mm}C{4mm} c C{4mm}C{4mm} c C{4mm}C{4mm} c C{4mm}C{4mm} c C{4mm}C{4mm}}
    \toprule
    && \multicolumn{2}{c}{Col. Hist} && \multicolumn{2}{c}{HAM10000} && \multicolumn{2}{c}{CB55} &&    \multicolumn{2}{c}{Flowers} &&    \multicolumn{2}{c}{ImageNet}  &&    \multicolumn{2}{c}{GPDS} \\
    \cmidrule{3-4} \cmidrule{6-7} \cmidrule{9-10} \cmidrule{12-13} \cmidrule{15-16}   \cmidrule{18-19} 
    \#Param  & Model & \% & $\sigma$ && \% & $\sigma$ && \% & $\sigma$ && \% & $\sigma$ && \% & $\sigma$ && \% & $\sigma$ \\
    \midrule
    \multicolumn{2}{l}{\quad \quad \quad Baselines} \\
    132K & BaselineConv                  &         83.2 & 6.5  &&         61.4  & 2.7 &&         90.0  & 0.3 &&         65.1  & 2.2 &&     26.9 & 0.1       &&   82.4 & 3.2 \\
    141K & BaselineDeep                  &         82.1 & 4.5  &&         58.4  & 3.5 &&         90.2  & 0.5 &&         63.4  & 2.6 && \textbf{28.9} & 0.3  &&   81.3 & 2.9 \\
    \multicolumn{2}{l}{\quad \quad \quad Matrix Transform} \\
    137K & RND Single                    &         86.2 & 1.5  &&         64.5  & 2.5 &&         97.3  & 0.4 &&         67.1  & 2.5 &&     27.2 & 0.1       &&   77.7 & 8.2 \\
    138K & RND Multi                     &         84.7 & 2.1  &&         73.4  & 0.9 &&         97.3  & 0.3 &&         60.0  & 9.5 &&     24.9 & 1.0       &&   83.2 & 2.3 \\
    \multicolumn{2}{l}{\quad \quad \quad Spectral} \\
    131K & DFT Single                    &         87.0  & 1.9 &&         63.4  & 3.0 &&         97.6  & 0.4 &&         68.1  & 3   &&     23.7 & 1.3       &&   66.6 & 4.1 \\
    135K & DFT Multi                     & \textbf{90.1} & 2   &&         70.8  & 4.7 && \textbf{97.7} & 0.2 &&         69.7  & 3   &&     22.0 & 0.6       &&   73.8 & 6.1 \\
    137K & DCT Single                    &         83.9  & 1.3 &&         66.4  & 3.4 &&         97.2  & 0.3 &&         66.7  & 4.7 &&     26.0 & 0.4       &&   81.8 & 2.8 \\
    138K & DCT Multi                     &         88.1  & 1   && \textbf{74.1} & 0.7 && \textbf{97.7} & 0.2 && \textbf{70.0} & 1.8 &&     22.5 & 0.3       &&   \textbf{86.2} & 2.5 \\
    \multicolumn{2}{l}{\quad \quad \quad References} \\
    57M  & AlexNet\cite{krizhevsky2012}  &         83.6  & 3   &&         66.3  & 2.7 &&         96.9  & 0.1 &&         71.1  & 3.2 &&     44.8 & 0.1       &&   92.0 & 1.2 \\
    11M  & ResNet-18\cite{he2016deep}    &         85.6  & 8.9 &&         76.4  & 0.6 &&         97.5  & 0.3 &&         70.8  & 4.9 &&     52.0 & 0.2       &&   98.0 & 0.3 \\
    \bottomrule
    \end{tabular}
    }
    \end{sc}
    \end{small}
    \end{center}
    \vskip -0.1in
\end{table*}


\definecolor{customViolet}{HTML}{7570b3}
\definecolor{customPink}{HTML}{e7298a}
\definecolor{customGreen}{HTML}{1b9e77}
\definecolor{customOrange}{HTML}{d95f02}

\begin{figure}[!t]
\centering
\includegraphics[width=\columnwidth]{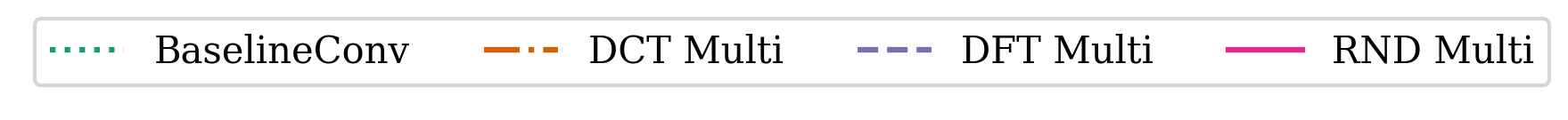}

\subfloat[ColorectalHist]{
\includegraphics[width=0.47\columnwidth]{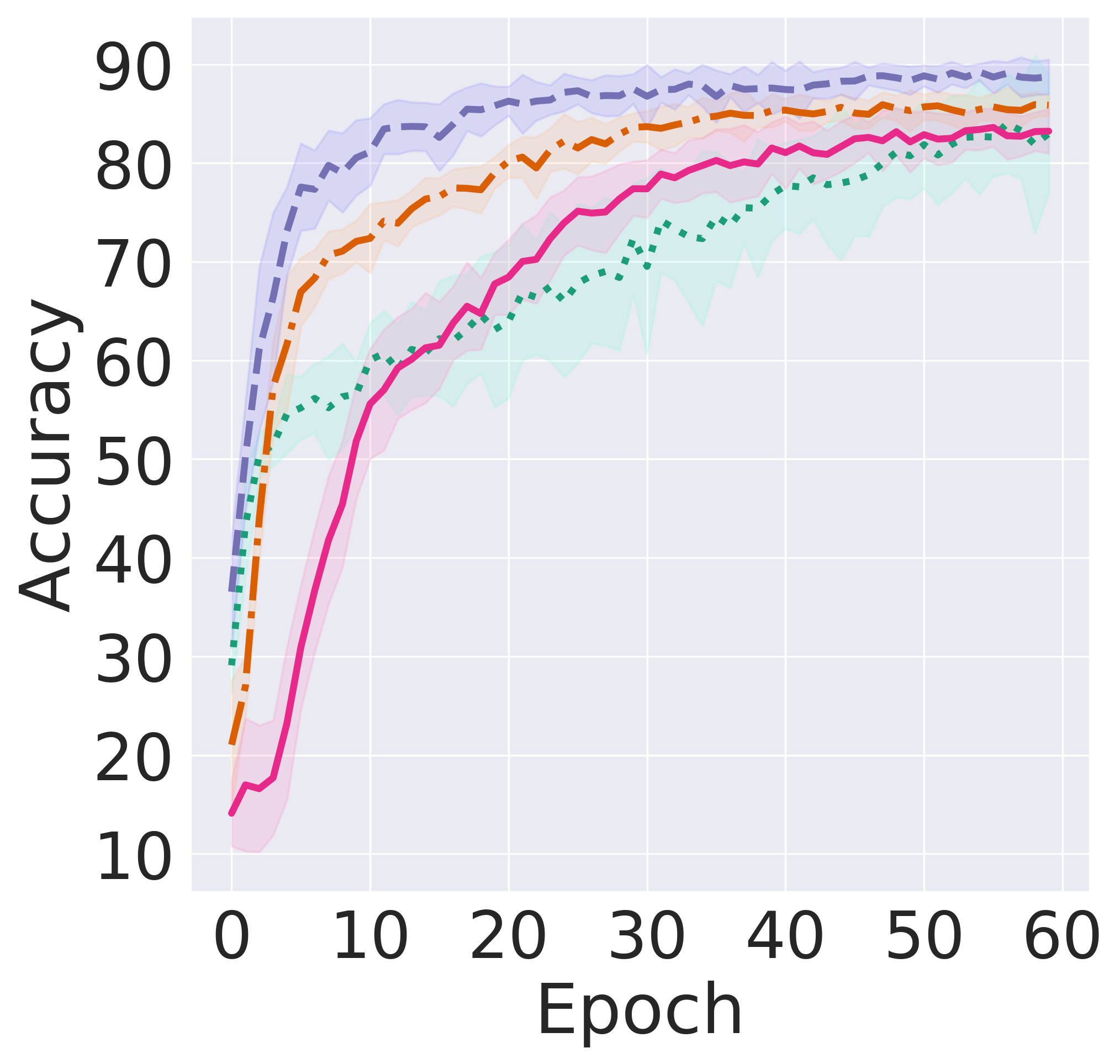}\label{subfig:convergeance_colorectal}}
\hfil
\subfloat[HAM10000]{
\includegraphics[width=0.47\columnwidth]{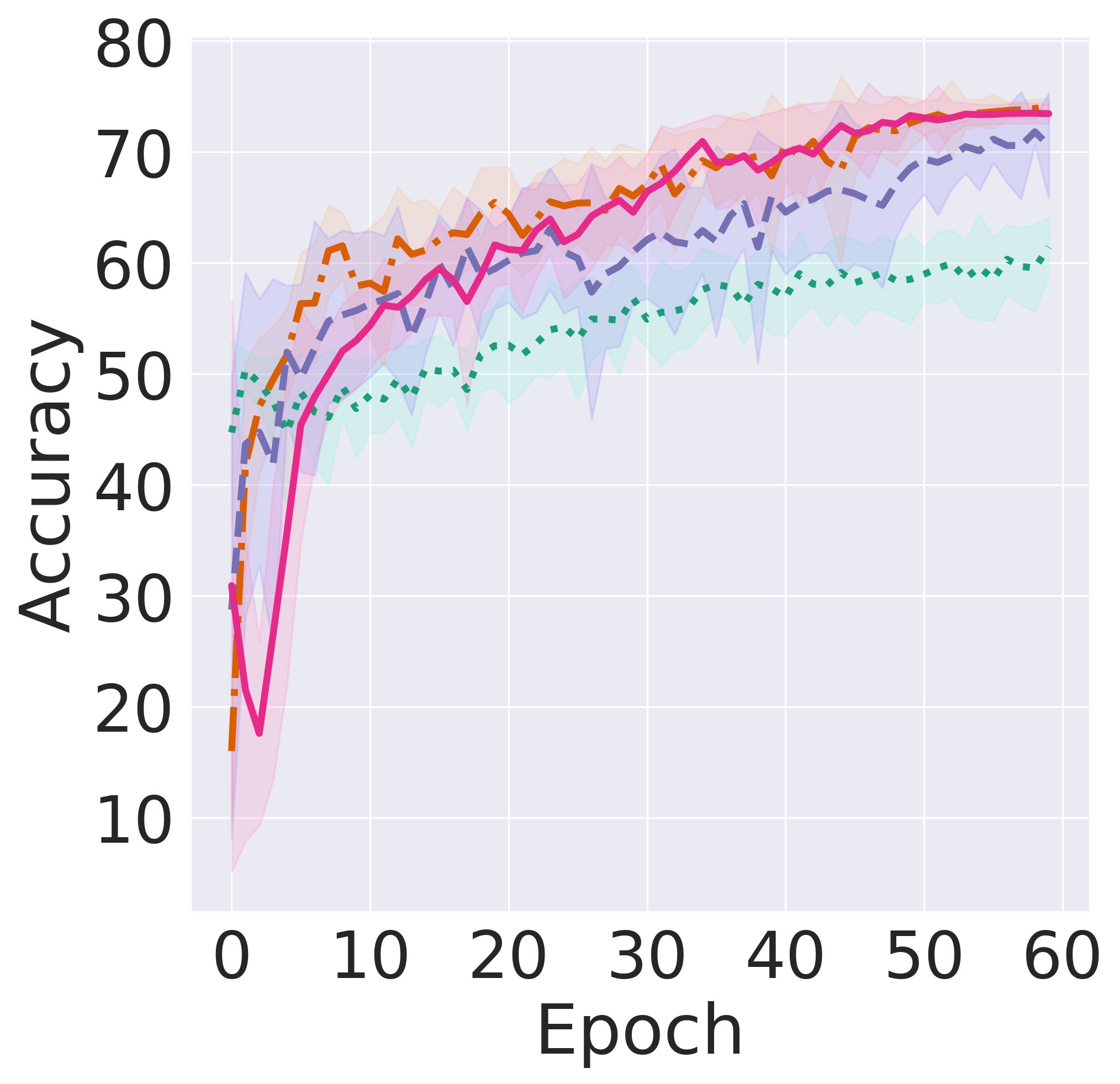}}
\vfil
\subfloat[CB55]{
\includegraphics[width=0.47\columnwidth]{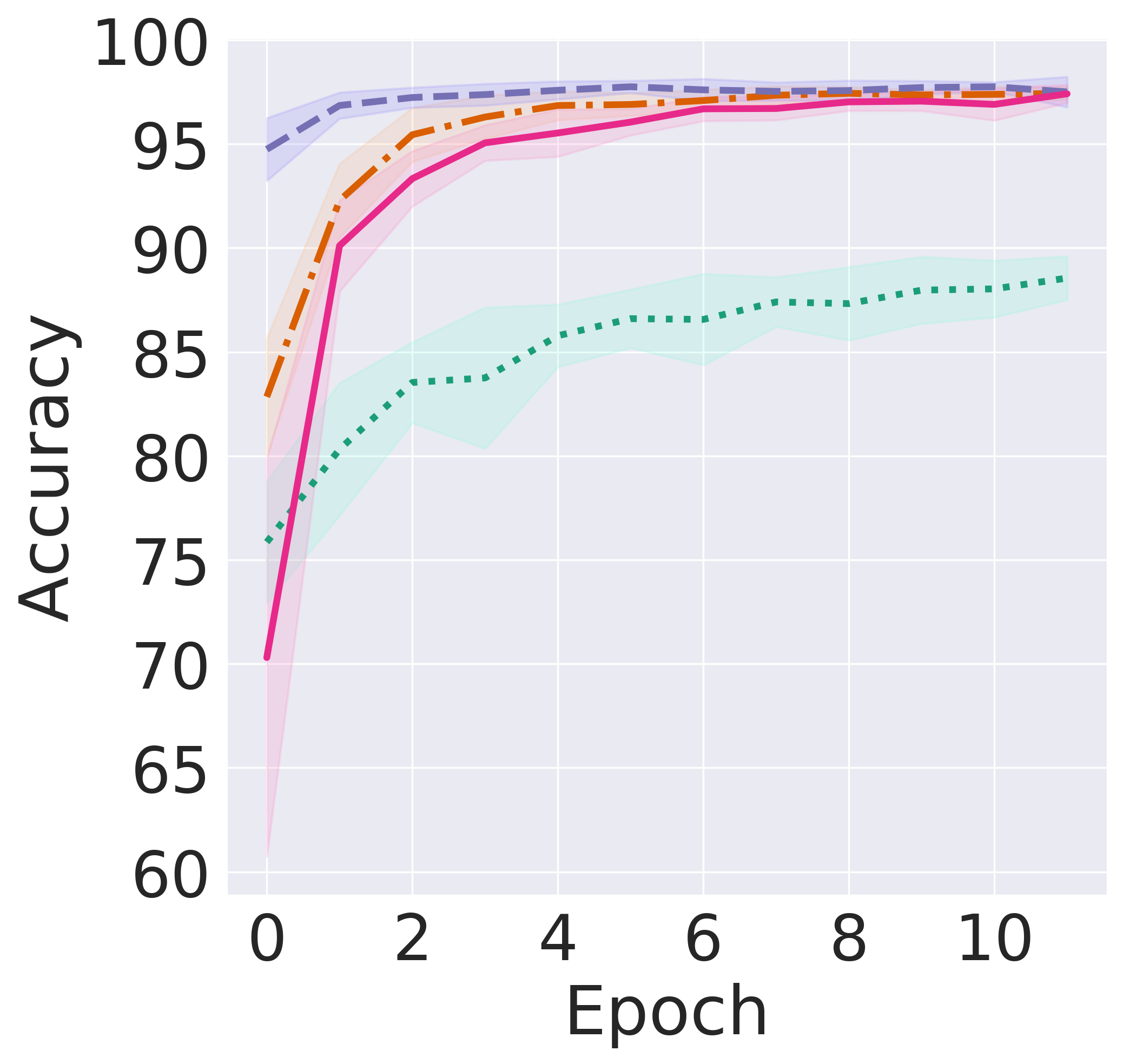}}
\hfil
\subfloat[Flowers]{
\includegraphics[width=0.47\columnwidth]{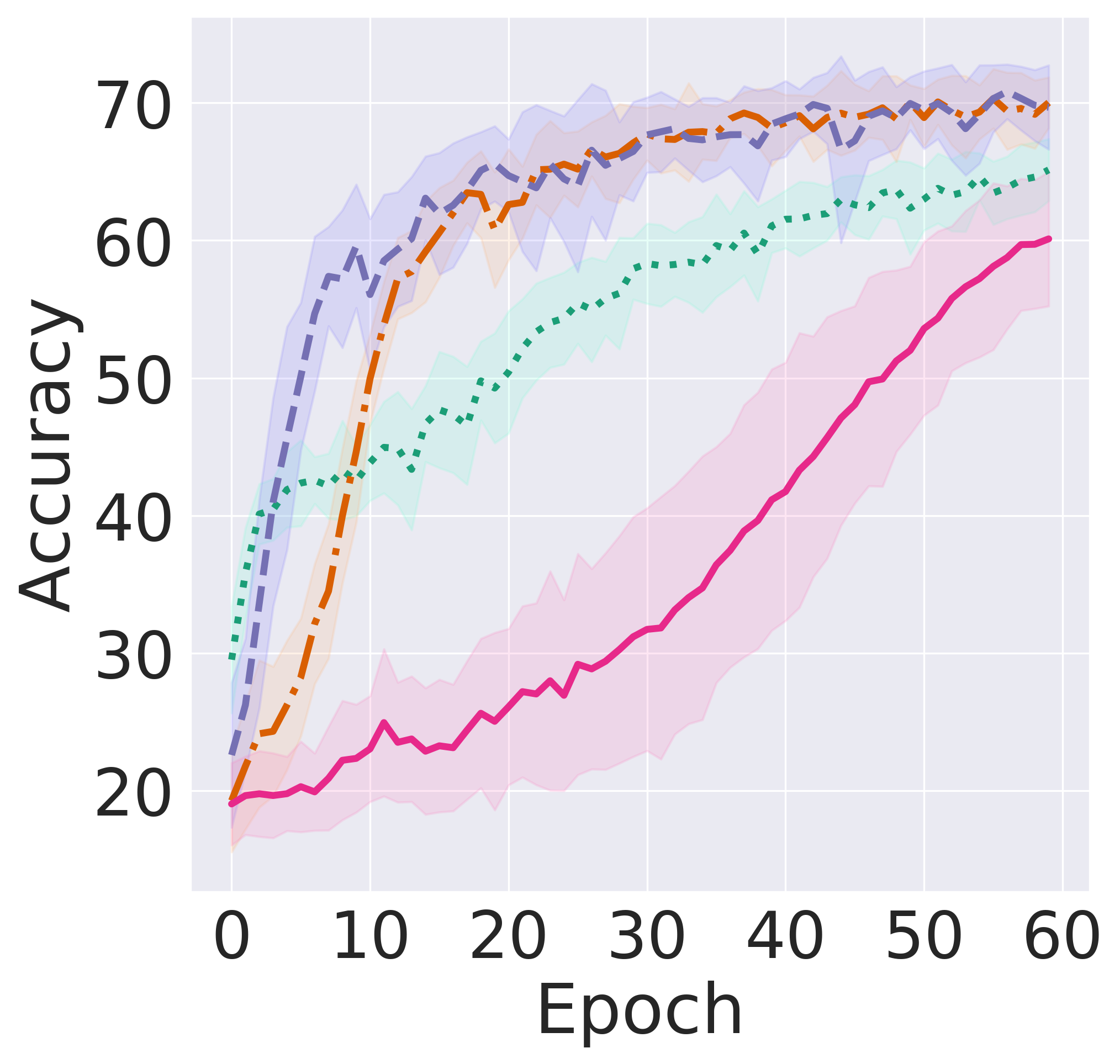}\label{subfig:convergeance_flower}}
\vfil
\subfloat[GPDS]{
\includegraphics[width=0.47\columnwidth]{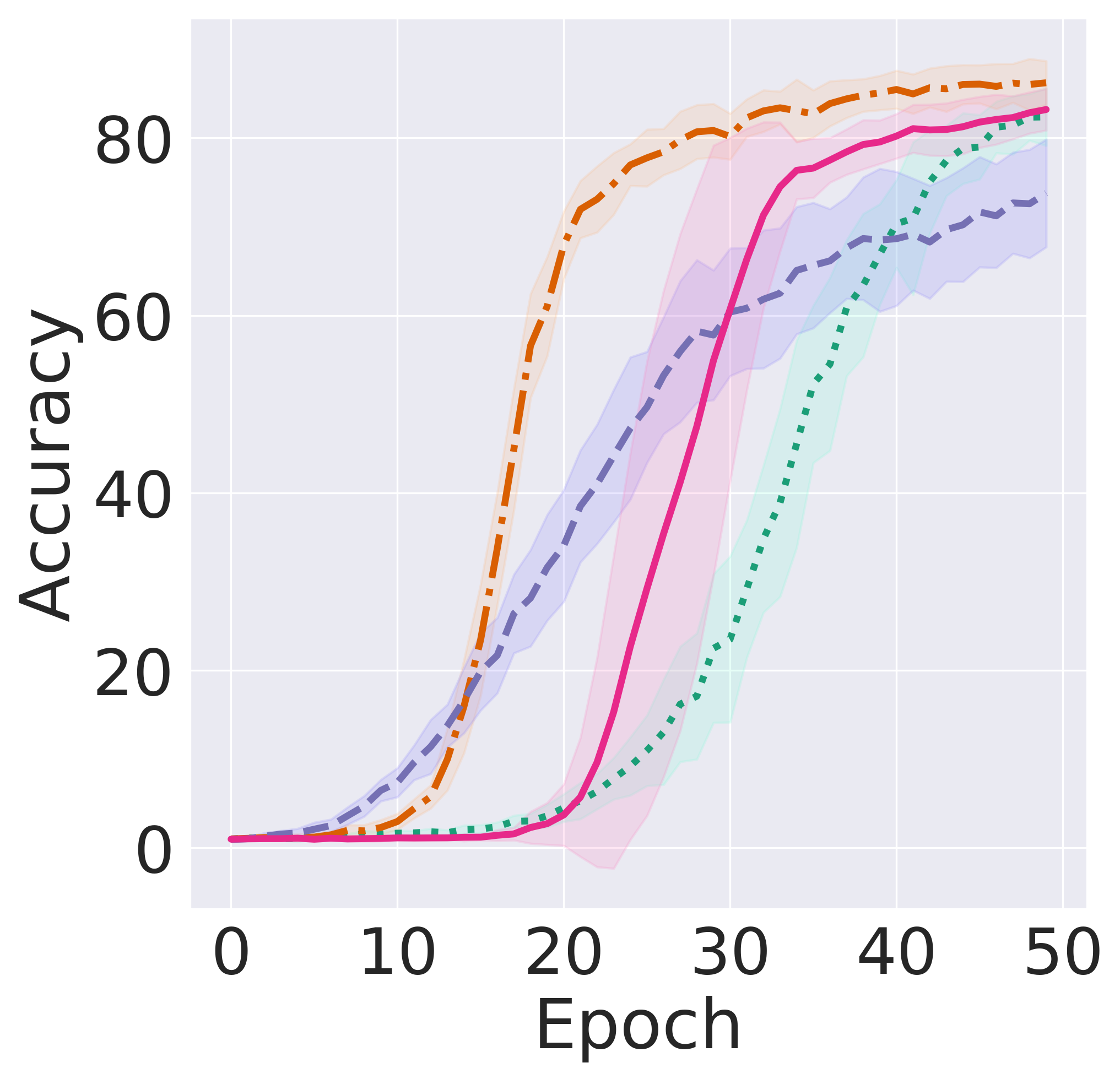}}
\hfil
\subfloat[ImageNet]{
\includegraphics[width=0.47\columnwidth]{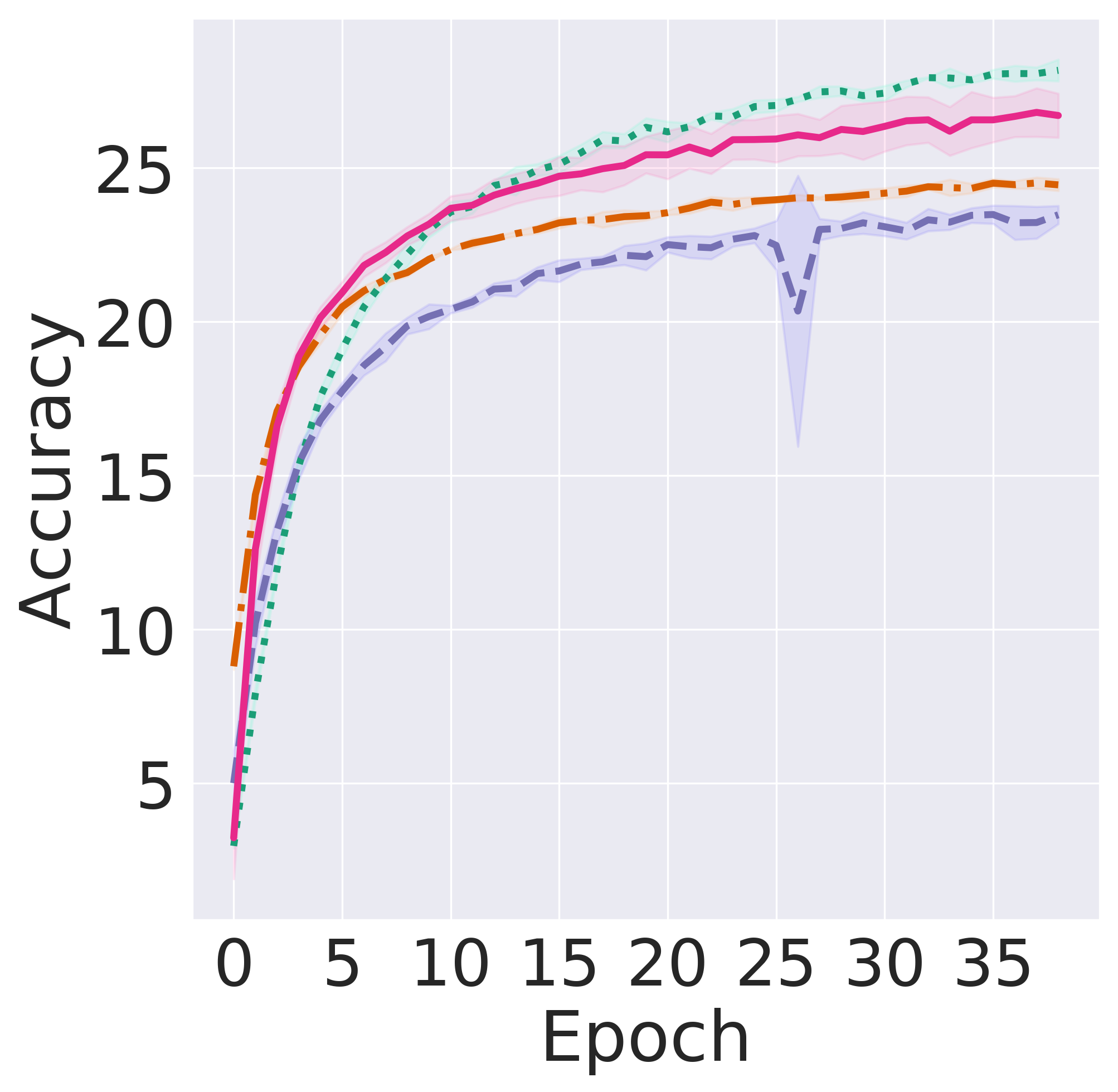}}

\caption{%
Visualization of validation accuracy of the four selected models \textcolor{customGreen}{BaselineConv}, \textcolor{customOrange}{DCT Multi}, \textcolor{customViolet}{DFT Multi} and \textcolor{customPink}{RND Multi} on the six datasets.
The data shown is collected over 20 distinct runs.
The different model configurations are explained in Section~\ref{toc:models_configuration}. 
The spectrally initialized configuration is characterized by a faster convergence and a smaller variance in the first phases of the training routine. 
Note that there are adjusted scales on each plot concerning the y-axis.
}
\label{fig:convergence}
\end{figure}


\begin{figure*}[!t]
\centering

\subfloat[DCT trained]{
\includegraphics[width=0.32\textwidth]{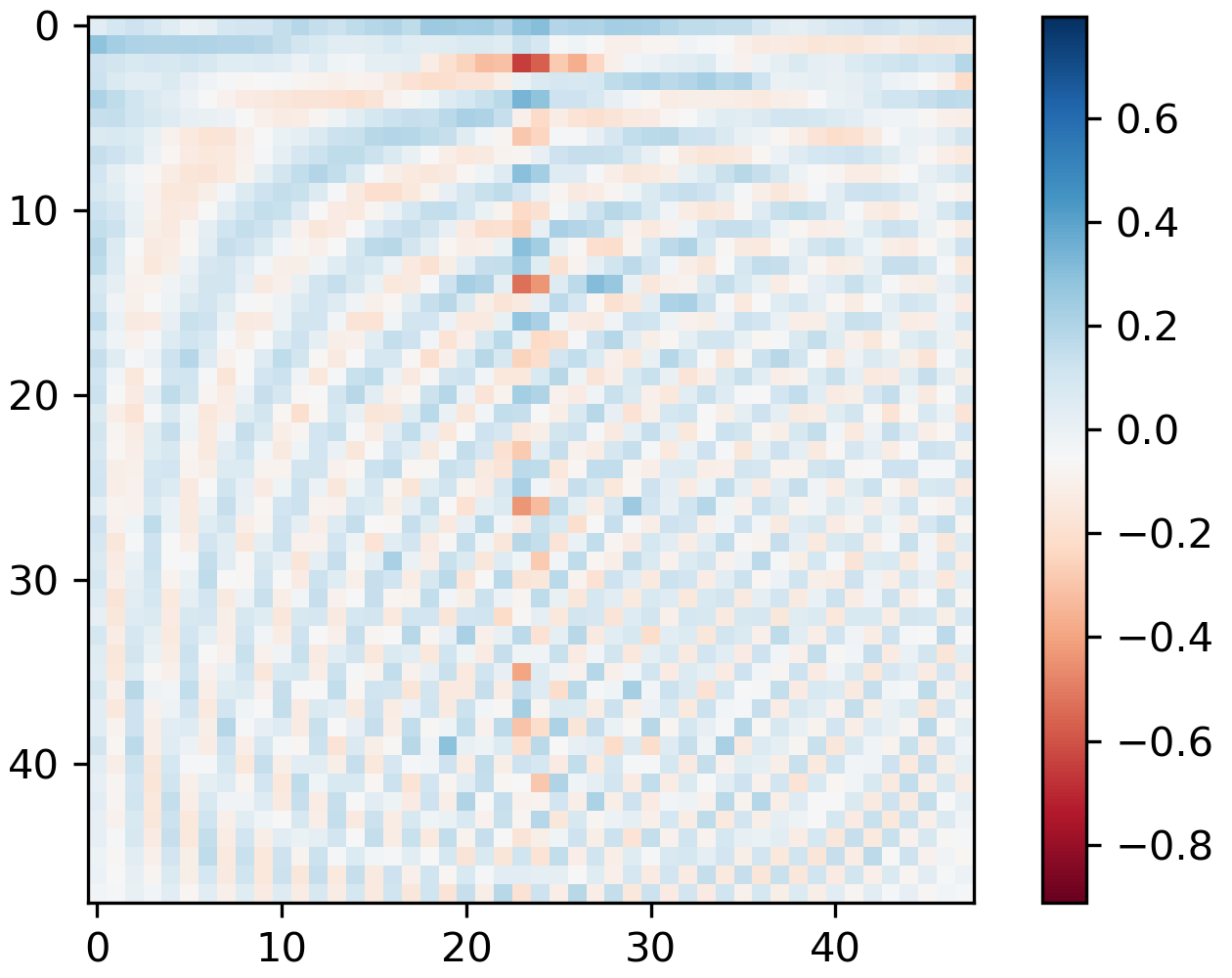}}
\hfil
\subfloat[DFT trained]{
\includegraphics[width=0.32\textwidth]{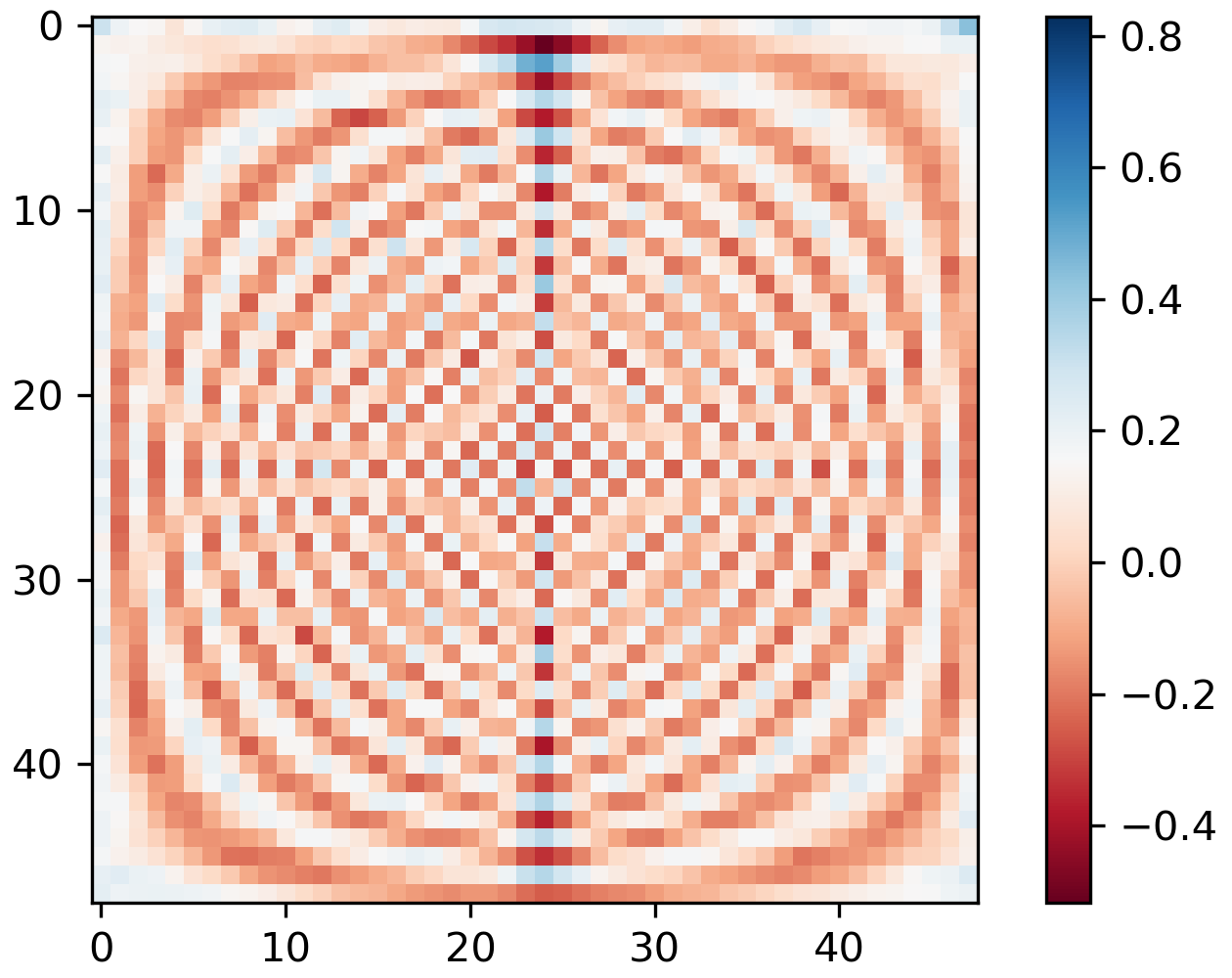}}
\hfil
\subfloat[RND trained]{
\includegraphics[width=0.32\textwidth]{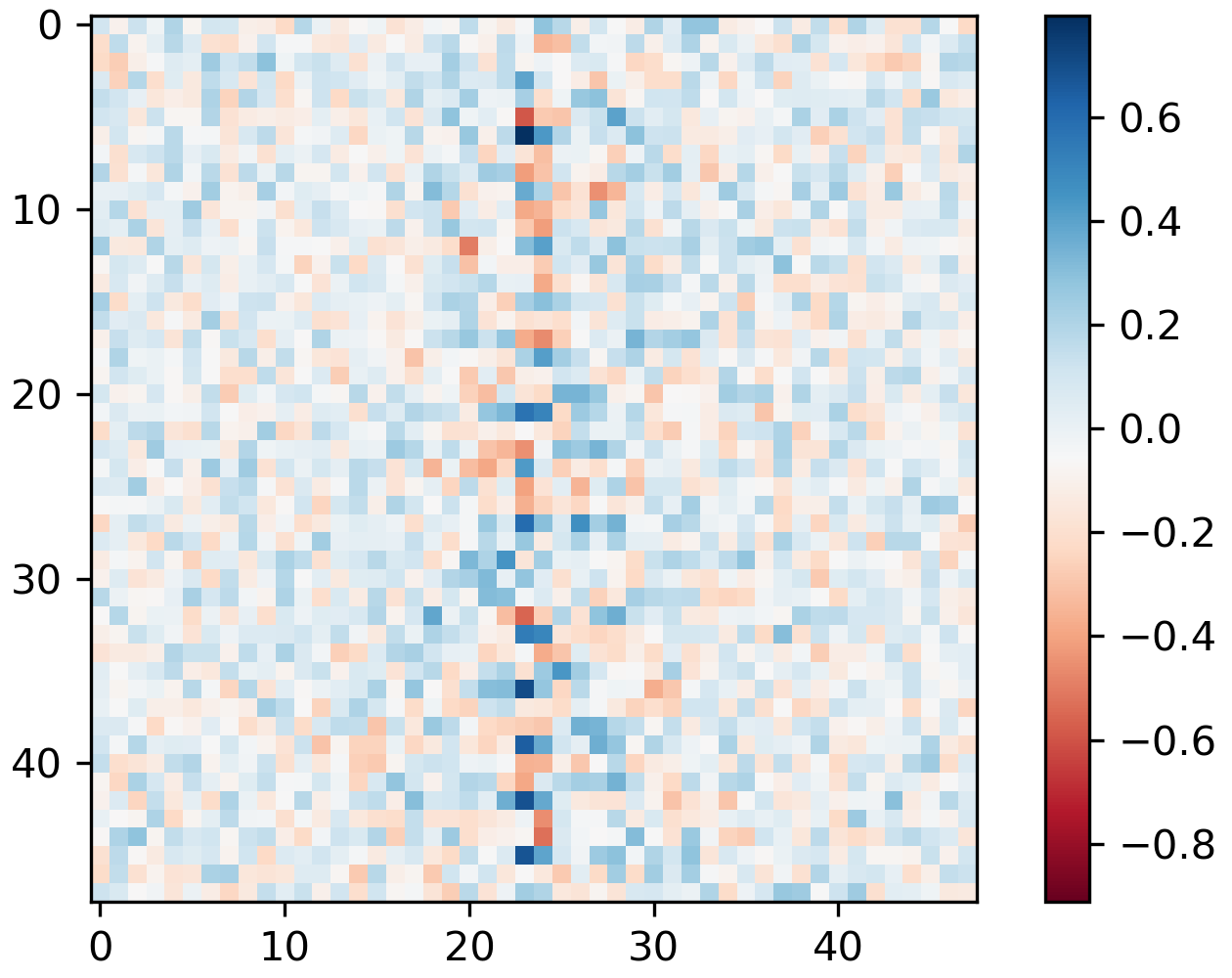}}

\caption{%
Visualization of the first weight matrix of a matrix transform used for training.
All three transforms are trained on the CB55 dataset, which features historical document classification based on the central pixel (thus, the central pixel decides on the class). 
The weight matrix from all three transforms shows a distinct vertical line on top of the remainder of the initialized transform.
It is visible by this emerging vertical pattern that the transform is adjusted to this special, globally present classification scenario.
} 
\label{fig:filters}
\end{figure*}

\subsection{Performance Comparison}\label{toc:PerformanceComparison}
The results of the different models on the datasets are shown in Table~\ref{tab:results}.
Throughout all datasets, the models with added matrix transformations perform overall better than the baseline model, with a mean gain of 2.2\% accuracy.
However, the performances differ greatly across specific datasets and initialization techniques.
The performance gains range from an average of 7.5\% on CB55 to 1.8\% on Flower, with a single peak of 12.7\% on HAM10000.
The performance loss spans from an average of -2.5\% on ImageNet to -4.1\% on GPDS, with a single peak of -15.8\% on GPDS\footnote{Noteworthy, the DFT spectral initialized models show relatively poor performances on GPDS, whereas the DCT outperforms the baseline.
While we are unsure of the exact causes of these lower performances, we believe that the Fourier transform does not capture the particular properties of the input data on signatures (black text on white background, always centered in the image) as well as the discrete Cosine one.}.

These results suggest that the structure of the dataset is essential for the choice of the architectural component, e.g., the selection of this new component.
Depending on what information is relevant to the task at hand, changing the representation of the data from spatial to the frequency domain is more or less beneficial.
Evidence suggests that datasets where the relevant classification information is uniformly distributed over the image, i.e., tissue classification and other pattern-heavy problems, profit more from our approach, than sets, where the relevant information is centered on a few closely distributed pixels.
Moreover, datasets that exhibit repeating structures or patterns seem to benefit the most from the spectral layer.
This hypothesis is supported by empirical evidence e.g., on Colorectral Hist. where the spectral initialized network (DFTMulti) surpasses the reference networks (AlexNet and ResNet) by 6.5\% and 4.5\%, respectively. 
This is promising considering the large gap in terms of the number of parameters (and architectural advances) featured in these models.
This trend is reinforced by the other medical dataset (HAM10000), the historical one (CB55), and in smaller magnitude by the natural images (Flower) one. 

In contrast, we expect datasets like ImageNet to benefit less from using the matrix transformations.
In fact, the information distribution within an image of such datasets is not denoting the important properties aforementioned. 
For example, border pixels contribute less than more central pixels in object classification datasets because of the bias of the photographer to center the subjects being captured. 

\subsection{Regularization Properties}
\label{toc:regularization_properties}
To asses whether the additional performance is due to a regularizing effect introduced by the non-separability of the matrix multiplication operation inside a module, we run the same experiments with another baseline model (in Table~\ref{tab:results} referred as ``BaselineDeep'') which has an additional convolutional layer with subsequent non-linearity. 
The results, however, indicate that this is not the case. 
It is thus reasonable to assume that the introduced transformations themselves have a beneficial effect on convergence behavior (Figure~\ref{fig:convergence}) and accuracy (Table~\ref{tab:results}). 
The addition of these linear transformations allows our simple three layers model to match and often outperform AlexNet and ResNet-18 models~--~which possess orders of magnitude more parameters (420 and 81 times more, respectively) and improvements such as batch-normalization layers or skip connections.

\subsection{Analysis of the Learned Transformations}
\begin{figure}[!t]
\centering

\subfloat[Input patch]{
\includegraphics[width=0.47\columnwidth]{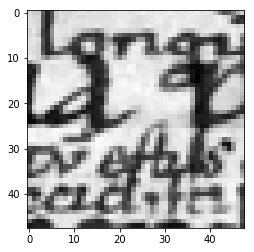}}\label{fig:patch_input}
\hfil
\subfloat[Feature map (init)]{
\includegraphics[width=0.47\columnwidth]{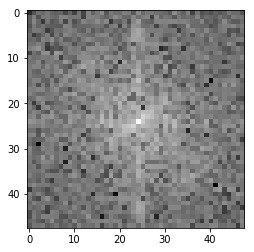}}\label{fig:patch_before}
\vfil
\subfloat[Feature map (learned)]{
\includegraphics[width=0.47\columnwidth]{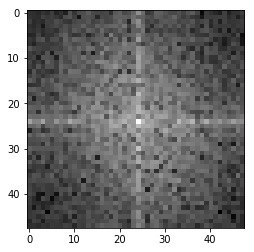}}\label{fig:patch_after}
\hfil
\subfloat[Normalized difference]{
\includegraphics[width=0.47\columnwidth]{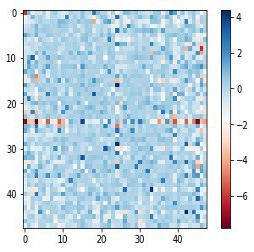}}\label{fig:patch_diff}

\caption{%
Visualization of the effect on the input patch (a) of the first DFT transform before (b) and after training (c). In (d), we report the normalized difference between (b) and (c). 
Visually, the difference from (b) to (c) is that the corners are much darker, which means the high mixed frequencies were lowered during training. 
This is also supported by the average of -2 seen in (d), which means the filters are overall losing energy due to the training. 
By looking closely at (c), one can see the $'+'$ pattern of increased values (brighter pixels), which means that the trained weights are looking for high frequencies along the horizontal and vertical axis but not diagonally (or another pattern). This seems to be a reasonable hypothesis since we are operating on text input data in this case. 
} 
\vspace{-3mm}
\label{fig:patch}
\end{figure}

While the observed convergence speedups of the DFT and DCT initialized transforms were expected and are in line with similar observations from other literature~\cite{SpectralConvolutionalNeuralNetworks, DeepFeatureDCT}, the respectable classification performance gains come rather unexpected.
We suspect that the transforms and especially the fact that they are trainable can be useful, allowing for the exploitation of the global input domain-specific structure. 
The potentially more meaningful and sparser representation produced by the DCT and DFT initialized transforms might also lead to smoother error surfaces, which are easier to optimize and decrease the chance of becoming stuck in a local optimum. 
To further investigate, in Figure~\ref{fig:filters}, we visualize the trained weights of the first transformation matrix with regards to the different weight initializations (DCT, DFT, RND). 
All three transforms are trained on the CB55 dataset, which features historical document classification based on the central pixel (thus, the central pixel decides on the class). 
It is visible by the emerging vertical pattern in all three matrices that the transform is adjusted to this particular, globally present classification scenario. 
Moreover, in Figure~\ref{fig:patch}, we visualize the effect that the first matrix module block has on the input patch (a): before~(b) and after~(c) training on CB55.
It seems that the weights of the higher mixed-frequencies become smaller in magnitude, which is supported by the darker corners in (c) as well as by the negative average of the normalized difference between before/after training (d).
Further, visually, there is a central $+$ pattern, which becomes more prominent after training (b,c). 
This is, in our opinion, a reasonable effect as we are operating on text input data.

\subsection{Single vs Multiple Matrix Transforms}
Adding multiple transforms (one after each convolutional operation) leads to significant performance gains - especially for the spectral initialization - as shown in Table~\ref{tab:results}. 
In fact, the ``Multi'' configurations with spectral initialization (\ac{DFT}/\ac{DCT}) performs generally better than their ``Single'' counterpart. 
The magnitude of the margin varies from almost 10\% on HAM10000 to a very marginal one on CB55 and even negative on ImageNet.
These results are aligned with the work done in~\cite{moczulski_acdc:_2016}, where a combination of specially structured linear layers, featuring both a forward and backward Cosine transformation, is shown to have good performances.
In the same manner, our ``Multi'' configuration seems to be able to leverage this ``forth-and-back'' transformations to some extent.
Allowing the network to work on the data both in the normal and spectral domains seems to be highly beneficial.

In the case of the random initialization, there are two cases in which the gain is negative, specifically on ColorectalHist and Flowers.
This is, to some extent, surprising because RND Multi outperforms RND Single by roughly 9\,\% on HAM10000, and exhibits similar behavior as the spectral counterpart on CB55.
Moreover, by closer inspection, we observe that the convergence rate of RND Multi is appreciably slower than its single counterpart on the Flowers dataset, as can be seen in Figure~\ref{subfig:convergeance_flower} and~\ref{subfig:convergeance_colorectal}.  
Our tentative explanation is that the larger amount of dissonant degree of freedom in the later stages of the network (as opposed to the spectral initialized ones) dampens the magnitude of the relative error, which gets back-propagated to the lower layers.
We are, however, unsure about the reason why this phenomenon is not observed on all datasets in the same strength.

\subsection{Comparison of Different Initialization}
Classification performance of differently initialized matrix transformations on the feature maps of CNNs appear to be mostly dataset dependent.
For example, \ac{DFT} initialization has better performance on ColorectalHist but loses against the \ac{DCT} initialized counterpart on HAM10000.
On CB55 and Flowers, both spectral initialization, show similar results.
However, it seems that the dataset CB55 exhibits a common ground around 97\,\% accuracy as all models, but the baselines achieve a similar result.
Regarding the spectral vs. random initialization, overall, the two unitary spectral transforms (DCT and DFT) do perform better, with an average gap of 1.6\,\% between the best spectral model and the best RND model. 
Even so, the main difference can be seen in convergence behavior.
Here, models containing the randomly initialized transform tend to converge significantly slower, as can be seen in Figure~\ref{fig:convergence}.

\subsection{Limitations and Outlook}
The obtained results are very promising and the final framework is easy to implement and use.
The most prominent drawback of using the proposed transformation seems to be the need to asses whether or not the dataset shows patterns that are suitable to be highlighted by such transformations. 
While this can generally be estimated by analyzing the visual appearance of images, a more analytical tool to support this decision would be very helpful. 
A promising direction would be the approach presented in~\cite{fleet_visualizing_2014}.
As discussed in section~\ref{toc:PerformanceComparison}, datasets with uniformly distributed classification information, such as tissue, profit from spectral transformations.
When measuring the occlusion sensitivity of the images in a trained CNN-network, we would expect to have a high probability of correct classification, regardless of which part of the image is covered.
On the other hand, datasets like ImageNet are more sensitive to this method depending on the omitted part of the image, as shown in~\cite{fleet_visualizing_2014}.
This coincides with the datasets our network-structures are not performing well on.
This might be used as a measurement for well-suited or ill-suited datasets.

In the case of the spectrally initialized transforms, it could also be promising to adapt spectral pooling operations ~\cite{SpectralConvolutionalNeuralNetworks}, potentially further exploiting the more compact spectral representations.

In this work, we constrained our investigations into the image domain, while this framework could be applied to higher-dimensional data as well. 
Further research should also focus on the impact of using spectrally initialized matrix transforms in neural networks applied to tasks such as video or medical tomography data analysis.
Overall, our results suggest that transforms, such as the spectral ones, are an often neglected part in image classification tasks.
It is commonly assumed that CNNs do not need the input to be transformed, probably because standard benchmark datasets reflect tasks where the original spatial representation is already ideal or close to it.
However, as we found, less mainstream image domains can benefit from transforms, even more so if the transforms are trainable.

%% file: sections/conclusion.tex
\section{Conclusion}
\label{toc:conclusion}
This work introduces trainable matrix transformations that are applied on feature maps and can be initialized with two unitary transforms, namely the \ac{DCT} and \ac{DFT} (and their respective inversions).
Experiments on six challenging datasets show that the transforms can exploit global input domain structure, resulting in significantly better results when compared to a similar baseline model, with an average performance gain of 2.2\% accuracy.
We further show that the spectral initialization as \ac{DCT} or \ac{DFT} brings substantial speedups in terms of convergence, when compared to random initialization.

All of the source code, documentation, and experimental setups are realized in the DeepDIVA framework and made publicly available in GitHub repositories.
The methods proposed in this paper lay out the basis for the realization of trainable spectral transformations and can be extended to higher-dimensional data as well.

%% file: sections/appendices.tex
\begin{appendices}

\section{Layer Implementation}
\label{app:layerImplementation}

The \acf{DFT} as defined in~\eqref{fourier2DM} returns a complex signal. As of November 2018, \emph{PyTorch} can not handle complex numbers directly~\footnote{\url{https://github.com/pytorch/pytorch/issues/755}}. 
Under the assumption that input signals are all real valued, the \acs{DFT} can be separated into real and imaginary part of the Fourier coefficients using Euler's formula
\begin{align*}
    e^{j\cdot x} = \cos{x} + j \cdot \sin{x}. 
\end{align*}
For a given Fourier weight matrix $W_K$ applying Euler's formula leads to
\begin{align*}
    W_K[k,n] &= e^{-2\pi j \frac{k\cdot n}{N}} \\
    & = \cos\left(-2\pi \frac{k\cdot n}{N} \right) + j \sin \left(-2\pi \frac{k\cdot n}{N} \right) \\
    & = W_{RK} + jW_{IK}
\end{align*}
Here, $W_{RK}$ denotes the real part and $W_{IK}$ the imaginary part of $W_K$.
This is done for all entries of the weight matrix in~\eqref{fourier2DM}
\begin{align*}
    W_{K} =&~ W_{RK} + jW_{IK} \\
    W_{L} =&~ W_{RL} + jW_{IL}
\end{align*}
The \ac{DFT} can then be computed as 
\begin{align*}
    X  = &~ (W_{RK} + jW_{IK})\cdot x\cdot (W_{RL} + jW_{IL})^T \\
    \begin{split}
     = &~  (W_{RK}xW^T_{IL} - W_{RK}xW^T_{IL}) \\
     & + j (W_{IK}xW^T_{RL}+W_{RK}xW^T_{IL}) \\
     = &~ X_R + jX_I
    \end{split}
\end{align*}
Effectively, the \ac{DFT} has twice as many parameters as the \ac{DCT}.
The \ac{DFT} does not contain more information than the \ac{DCT}, the surplus of parameters is due to symmetry properties of the \ac{DFT} if the input is real valued. 
This is discussed in appendix~\ref{app:redundancy}.
Our implementation has the option to handle the \ac{DFT} output in two different ways.
Either the output consists of real part $X_R$ and imaginary part $X_I$ of the transformation. 
These two are concatenated, effectively doubling the size of the feature map, i.e., the output is $\hat{X} = [X_R^T, X_I^T]^T$.
The other way is the computation of the amplitude.
In this case, the amplitude of the output signal is computed as is usually done when working with frequency analysis in signal processing, i.e., the output is $\hat{X}[k,l] = \sqrt{X_R^2[k,l] + X_I^2[k,l]}$.
The resulting feature map has same dimension as the input.

In our evaluations, the complex valued output is used.

\section{Redundancy of data}
\label{app:redundancy}

For a real-valued input signal $x \in \mathbb{R}^{N}$, its Fourier transform $X \in \mathbb{C}^N$ has important symmetry properties.
For even $N$, $X_k =  X^*_{N-k}$ and $X_0, X_{\frac{N}{2}} \in \mathbb{R}$. For odd $N$,  $X_k =  X^*_{N-k}$ and $X_0 \in \mathbb{R}$. 
Where $X_i^*$ denotes the complex conjugate of $X_i$.
As a result, the \acs{DFT} of even (respectively odd) length real input signals are defined by the first $\frac{N}{2}+1$ (respectively $\frac{N+1}{2}$) entries of the transformed signal. 
The remaining $\frac{N}{2}-1$ (respectively $\frac{N-1}{2}$) entries can be computed from the existing ones. 
Or in other words, the degree of freedom for the \acs{DFT} of an even (respectively odd) real input signal is $\frac{N}{2}+1$ (respectively $\frac{N+1}{2}$). 

If the input signal is complex valued, then there are no more redundancies, but the layer implementation as explained in Appendix~\ref{app:layerImplementation} would have to be expanded by a complex valued input $x = x_R + j x_I$.

While this redundancy of parameters is a reduction of memory, this can pose a problem for the back transformation of the \acs{DFT}. For a \acs{DFT} signal to be back transformable to a real valued signal, these properties of symmetry need to be fulfilled. Consequently, about half of the parameters need to be fixed to meet this requirement. This is also discussed in~\cite{SpectralConvolutionalNeuralNetworks}. So far, in our implementation, this problem of redundancy is not taken care of in the unfixed layer case.

\section{Back transformation}
\label{app:backtransform}

Back transformations of both \acs{DFT} and \acs{DCT} can be implemented in a similar manner to the forward transformation.
As the DFT matrices are unitary, the definition of the inverse matrices is straight forward. The inverse of a unitary matrix $W_u$ is its conjugate transpose $W_u^*$.
Alternatively, the definition for the backward transformation can be used to derive the back-transformation matrices
\begin{align*}
    x[n,m] = & \sum_{k=0}^{K-1} \sum_{l=0}^{L-1} x[k,l] e^{2\pi j \frac{kn}{N}} e^{2\pi j \frac{lm}{M}}\\
    x = & W_K^* \cdot X \cdot (W_L^T)^*, \\ 
    & W_K^*[n,k] = e^{2\pi j \frac{k\cdot n}{N}},~~ W_L^*[m,l] = e^{2\pi j \frac{l\cdot m}{M}}
\end{align*}
Using again Euler's formula, these can be expanded and implemented as described in Appendix~\ref{app:layerImplementation}.

The inverse of the \acs{DCT}-II is the \acs{DCT}-III multiplied by a factor $\nicefrac{2}{NM}$. This factor is merely for normalization and can be omitted or can be added to both forward and backward transformation equally as $\sqrt{\nicefrac{2}{NM}}$
\begin{align*}
    x =& \hat{W}_K \cdot X \cdot (\hat{W}_L)^T\\
    \hat{W}_K[n,0] =& \frac{1}{N}, ~~ \forall n  \\
    \hat{W}_L[m,0] =& \frac{1}{M}, ~~ \forall m  \\
    \hat{W}_K[n,k] =& \frac{2}{N}cos\left(\frac{\pi}{N}k(n+\frac12)\right), ~~~~ \forall n,~ \forall k > 0 \\
    \hat{W}_L[m,l] =& \frac{2}{M}cos\left(\frac{\pi}{M}l(m+\frac12)\right), ~~~~ \forall m,~ \forall l > 0
\end{align*}

\section{Linear layers and matrices}
\label{app:matrixdimensions}

\subsection{Linear layers in MLP} \label{app:MLPlinearlayer}
In a \acs{MLP}, linear layers are represented by matrix-vector multiplications $f: x \mapsto y=Wx$ with $x \in \mathbb{R}^{N}$ and $W\in\mathbb{R}^{M\times N}$. The weight matrix $W$ fully connects two different layers $x\in \mathbb{R}^N$ and $y\in\mathbb{R}^M$. There are $N\cdot M$ parameters in the weight matrix $W$.

The 2D-\acs{DFT} and 2D-\acs{DCT} are linear matrix-matrix multiplications as in equation~\eqref{fourier2DM} and~\eqref{dct2DM}.
They consist of linear operations on matrices, $W\cdot X$ usually with $W,X \in \mathbb{R}^{N\times N}$. There are $N^2$ parameters in the weight matrix $W$ for an input $X$ of dimension $N\times N$.
If the input is flattened row first, $X \in \mathbb{R}^{N\times N} \rightarrow x \in \mathbb{R}^{N^2}$, the weight matrix $W$ is adjusted to a sparse matrix $W \in \mathbb{R}^{N\times N} \rightarrow \tilde{W}\in \mathbb{R}^{N^2 \times N^2}$. The new weight matrix $\tilde{W}$ is a block matrix
\begin{align}
    \tilde{W} &=
    \left[\begin{array}{ccc}
         \tilde{W}_{11}& \hdots & \tilde{W}_{1N}\\
         \vdots& \ddots & \vdots \\
         \tilde{W}_{N1}& \hdots & \tilde{W}_{NN}
    \end{array} \right] \\
    \tilde{W}_{ij} &= \mathbb{I}_Nw_{ij}
\end{align}
with $\mathbb{I}_N$ the identity matrix of size $N$ and $w_{ij} = W[i,j]$ the entries of the original matrix. The matrix has density $\frac{1}{N}$, i.e., sparsity $\frac{N-1}{N}$. This is comparable to a not-fully connected linear layer with shared weights. 

\subsection{Kronecker Product}\label{app:kronecker}
For two matrices $A: [m \times n]$ and $B: [k \times l]$, the Kronecker product $A \otimes B$ is a $[mk \times nl]$ block matrix:
\begin{equation} \label{kroneckerprod}
    A \otimes B = 
    \left[\begin{array}{cccc}
         a_{11}B & a_{12}B & \cdots & a_{1n}B  \\
         a_{21}B & a_{22}B &        & a_{2n}B  \\
         \vdots  &         &        & \vdots   \\
         a_{m1}B & a_{m2}B & \cdots & a_{mn}B
    \end{array}\right]
\end{equation}
with $a_{ij}: A[i,j]$. 
The Kronecker product can be used for a more convenient representation of certain matrix equations. 
Given four matrices $A,B,C$ and $X$ of appropriate dimensions, the equation $A\cdot X\cdot B=C$ can be rewritten using the Kronecker product and matrix vectorization:
\begin{align}
\begin{split}
    A \cdot X \cdot B & = C \\
    & \Updownarrow \\
    (B^T \otimes A) \cdot vec(X) &= vec(C)
\end{split}
\end{align}
This allows the application of Gauss or other simple-to-use  algorithm to solve for $X$.\newline
If all matrices $A, B, X$ are of size $[n\times n]$, the complexity of multiplying the Kronecker product matrix $K = (B^T \otimes A)$ with $X$ is $\mathcal{O}(n^4)$ while the complexity of the original matrix multiplication $AXB$ is $\mathcal{O}(n^3)$. 
The two matrices $A$ and $B$ have a total of $2n^2$ parameters, while $K$ has $n^4$ parameters.  Rank information is maintained, $rank(K) = rank(A)rank(B)$.\newline
For the application to equation~\eqref{linmap2}, the Kronecker product can be used as
\begin{align}
    \tilde{g} &: U\! \rightarrow\! \tilde{V}, & U\! \subseteq\! \mathbb{R}^{N\times M}, \tilde{V}\!\subseteq\! \mathbb{R}^{K\cdot L \times 1},\\
    \tilde{g}(x) &: x \!\mapsto\! K \cdot vec(x), & K\! = \!(W_2 \otimes W_1). \label{kronecker}
\end{align}
Disadvantage of this method is the necessity to maintain the structure of the parameters of $K$ as described in~\eqref{kroneckerprod}. 
If all parameters of $K$ can be optimized independently, the original structure of $K$ can get lost and this layer will result in $n^4$ independent parameters instead of $2n^2$ independent parameters.
The resulting matrix will no longer represent $W_1$ and $W_2$, i.e. no longer be the Kronecker product.
Alternatively, if $rank(K)$ and the structure of the Kronecker product is to be maintained, there have to be additional constraints on the parameters, rendering the whole optimization even more complex.
In other words, the inherent block structure as described in equation~\eqref{kroneckerprod}, can not be maintained during training, resulting in a simple fully connected MLP layer as described in section~\ref{app:MLPlinearlayer}, but without the sparsity.
At the same time, computational complexity is increased and number of trainable parameters are increased.
The advantages of the Kronecker product can not be exploited in this scenario.
\section{Computing Network Number of parameters}
\label{toc:appendix_number_parameters}

In order to operate on image-based datasets, we design a \ac{CNN} model, shown in Figure~\ref{fig:model_architecture}.
The core structure is 3 module-block layers (explained in the next sections) followed by a \ac{GAP} and concluded by a fully connected classification layer.
In order to have comparable results, regardless of the configuration, each model has the same amount of parameters\footnote{With a $\pm3.5\%$ tolerance.} which is roughly 135k. 
Because of this, the size of the feature map or the number of filters for a specific convolution layer may vary across different configurations. 
The exact numbers are provided together with the source code, such that full reproducibility is guaranteed (details in Section~\ref{toc:reproducibility}).

\subsection{Baseline Module}

The baseline module is composed of a convolution layer followed by a Leaky ReLU activation function and is shown in Figure~\ref{fig:model_architecture}.
The number of parameters in this module is computed as $F_{w} \cdot F_{h} \cdot F_n \cdot I_d$, where $F_{w}$ and $F_{h}$ are the convolution filter sizes, $F_n$ is the number of convolutional filters and $I_d$ the input depth.

\subsection{Regular Module}

The regular module is similar to the baseline module with the addition of a spectral operation right after the convolutional layer, as shown in Figure~\ref{fig:model_architecture}.
This way a spectral transformation is performed on the features space and then subject to the activation function, as is common practice with neural networks. 
The number of parameters of this module is the sum of parameters of the convolutional layer and the spectral block. 
For the convolutional layer the procedure is identical as shown in the baseline module, whereas for the spectral block the number of parameters is $FM_w \cdot K + FM_h \cdot L$, with $FM_w$ and $FM_h$ being the width and height of the feature map after the convolutional layer, and $K,L$ is the number of frequencies examined by the spectral transformation.
In our experiments $K$ and $L$ are set to be equal to $FM_w$ and $FM_h$, therefore the final number of parameters for the spectral layer can be computed as $2 \cdot FM_{w}^{2}$ and $4 \cdot FM_{w}^{2}$ for the 2D-DCT and 2D-DFT layer, respectively.

\subsection{Global Average Pooling}

The Global Average Pooling has no parameters to be trained.

\subsection{Classification Layer}

The final classification layer is a simple fully connected layer and its number of parameters is $FM_w \cdot FM_h \cdot C$ with $FM_w$ and $FM_h$ being the width and height of the feature map after the \ac{GAP}, and $C$ the number of output classes on this particular dataset.

\end{appendices}